\documentclass[conference]{IEEEtran}
\IEEEoverridecommandlockouts
\usepackage{times}
\usepackage{epsfig}
\usepackage{graphicx}
\usepackage{subfigure}
\usepackage{diagbox}
\usepackage[skip=2pt]{caption}
\makeatletter
\newcommand*{\rom}[1]{\expandafter\@slowromancap\romannumeral #1@}
\makeatother

\newcolumntype{C}{>{\centering\arraybackslash} m{1.5cm} }
\makeatletter
\@namedef{ver@everyshi.sty}{}
\makeatother
\usepackage{multirow}
\usepackage[bookmarks=false]{hyperref}
\usepackage{cite}
\usepackage{amsmath,amssymb,amsfonts}
\usepackage{algorithmic}
\usepackage{textcomp}
\usepackage{xcolor}
\def\BibTeX{{\rm B\kern-.05em{\sc i\kern-.025em b}\kern-.08em
    T\kern-.1667em\lower.7ex\hbox{E}\kern-.125emX}}

\begin{document}

\title{Boosted GAN with Semantically Interpretable Information for Image Inpainting
}

\author{ 
\IEEEauthorblockN{1\textsuperscript{st} Ang Li}
\IEEEauthorblockA{
\textit{The University of Melbourne}\\
angl4@student.unimelb.edu.au}
\and
\IEEEauthorblockN{2\textsuperscript{nd} Jianzhong Qi}
\IEEEauthorblockA{
\textit{The University of Melbourne}\\
jianzhong.qi@unimelb.edu.au}
\and
\IEEEauthorblockN{3\textsuperscript{rd} Rui Zhang}
\IEEEauthorblockA{
\textit{The University of Melbourne}\\
rui.zhang@unimelb.edu.au}
\and
\IEEEauthorblockN{4\textsuperscript{th} Ramamohanarao Kotagiri}
\IEEEauthorblockA{
\textit{The University of Melbourne}\\
kotagiri@unimelb.edu.au}
}

\maketitle

\begin{abstract}
Image inpainting aims at restoring missing regions of corrupted images, which has many applications such as image restoration and object removal.
However, current GAN-based inpainting models fail to explicitly consider the semantic consistency between restored images and original images.
For example, given a male image with image region of one eye missing, current models may restore it with a female eye.
This is due to the ambiguity of GAN-based inpainting models: these models can generate many possible restorations given a missing region.
To address this limitation, our key insight is that semantically interpretable information (such as attribute and segmentation information) of input images (with missing regions) can provide essential guidance for the inpainting process.
Based on this insight, we propose a boosted GAN with semantically interpretable information for image inpainting that consists of an inpainting network and a discriminative network.
The inpainting network utilizes two auxiliary pretrained networks to discover the attribute and segmentation information of input images and incorporates them into the inpainting process to provide explicit semantic-level guidance.
The discriminative network adopts a multi-level design that can enforce regularizations not only on overall realness but also on attribute and segmentation consistency with the original images.
Experimental results show that our proposed model can preserve consistency on both attribute and segmentation level, and significantly outperforms the state-of-the-art models.
\end{abstract}


\section{Introduction}\label{secIntro}
\begin{figure}[htb]
\centering
\setlength{\tabcolsep}{0.2em}
    {
\begin{tabular}{cccc}

\includegraphics[width=1.9cm]{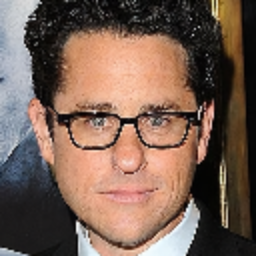}&
\includegraphics[width=1.9cm]{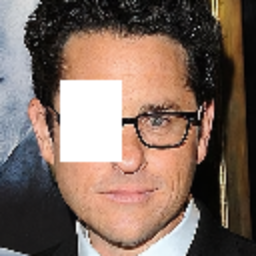}&
\includegraphics[width=1.9cm]{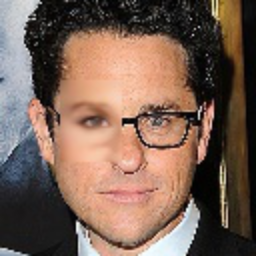}&
\includegraphics[width=1.9cm]{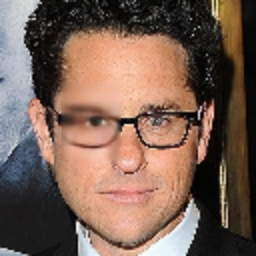}
\\

\includegraphics[width=1.9cm]{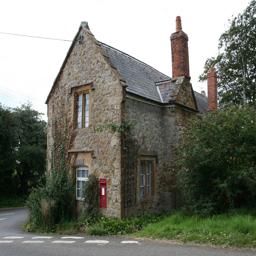}&
\includegraphics[width=1.9cm]{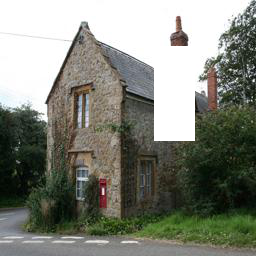}&
\includegraphics[width=1.9cm]{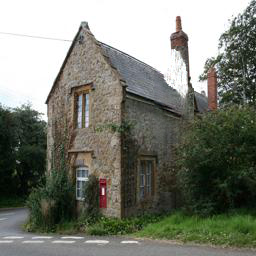}&
\includegraphics[width=1.9cm]{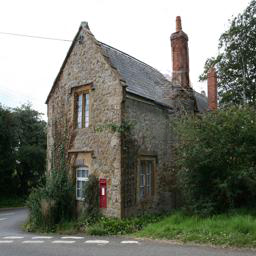}
\\

(a) Original & (b) Input & (c) GntIpt  & (d) Ours\\

\end{tabular}
}
\caption{Inpainting comparison with the state-of-the-art method GntIpt \cite{yu2018generative}.}
\label{exampleCompare} % I can do without the label too
\vspace{-0.8cm}
\end{figure}

Given an image where part of the image is missing, image inpainting aims to synthesize plausible contents that are coherent with non-missing regions.
Figure \ref{exampleCompare} illustrates the problem, where Figure 1b shows two input images where a region is missing from each original image.
The aim is to fill the missing regions, such that the filled regions contain contents that make the whole image look natural and undamaged (semantically consistent with the original images in Figure 1a).
With the help of image inpainting, applications such as restoring damaged images or removing blocking contents from images can be realized.

Traditional inpainting methods \cite{afonso2011augmented}\cite{barnes2009patchmatch}\cite{efros1999texture} mostly rely on a strong assumption that patches (or textures) similar to those in the missing regions should appear in the non-missing regions.
They fill the missing regions with background patches obtained from the non-missing regions.
These methods only focus on low-level features and thus fail to deal with situations where the missing regions contain non-repetitive contents.

Recently, \textit{Generative Adversarial Networks} (GAN) have shown a strong capability in image generation \cite{goodfellow2014generative}\cite{radford2016unsupervised}\cite{xiao2018}.
Inspired by the GAN model, image inpainting can be seen as a conditional image generation task given the context of input images \cite{yu2018generative}\cite{iizuka2017globally}\cite{pathak2016context}\cite{Yang_2017_CVPR}\cite{yeh2017semantic}.
A representative work following this model is \textit{Context Encoder} \cite{pathak2016context}, where a convolutional encoder-decoder network is trained with a combination of reconstruction loss and adversarial loss \cite{goodfellow2014generative} to predict the missing contents. 
The reconstruction loss guides to recover the overall coarse structure of the missing region, while the adversarial loss guides to choose a specific distribution from the results and promotes natural-looking patterns.

In spite of the encouraging results, the restored images by current GAN-based inpainting methods may be perceptually contradictory with their corresponding ground truth images.
%current GAN-based inpainting methods do not explicitly consider the image content consistency in terms of semantics.
%Specifically, the inpainting process may ignore semantic information of the non-missing context such as attribute information and segmentation information.
%Therefore, the restored images may be perceptually contradictory with the ground truth images. 
For example in Figure 1c, the inpainting results generated by the latest state-of-the-art inpainting method \textit{GntIpt} \cite{yu2018generative} encounter attribute mismatch (inconsistent on the attribute ``wearing eyeglasses" in the first image) and segmentation structure misalignment (part of the building missing in the second image) with the non-missing contexts.
Such limitations come from the ill-posed GAN-based formulation of the problem: the one-to-many mapping relationships between a missing region and its possible generated restorations.
This can be ascribed to the limitation of commonly used combination of loss functions: a reconstruction loss for pixel-wise identity and an adversarial loss for determining overall image realness.
The reconstruction loss mainly focuses on minimizing pixel-wise difference (very low-level), while the adversarial loss determines overall similarity with real images (very high-level).
Intuitively, mid-level (between the pixel and overall realness levels) regularizations which can provide more explicit guidances on semantics are needed.
This limitation can lead to sub-optimal inpainting results as shown in Figure \ref{exampleCompare} due to the lack of specific regularizations on semantic consistency in inpainting model training.

To address this limitation, our key insight is that semantically interpretable information of input images (with missing regions) can provide essential guidance for the inpainting process.
Here, \emph{semantically interpretable information} refers to human-defined label information that is understandable and has explicit meanings.
Arguably, the two most representative types of semantically interpretable information of images are image attribute information and image segmentation information:
%Specifically, we aim to utilize and incorporate two essential human-understandable information of input images: image attribute information (a vector representing multiple descriptive text labels related to an image, e.g., gender, hair color, and face shape of a facial image) and image segmentation information (a one-dimensional map that associates each pixel of an image with a region label, e.g., regions of eyes, mouth and nose from a facial image). 
(1) Attribute information is usually represented as a vector representing multiple descriptive text labels related to an image, e.g., gender, hair color, and face shape of a facial image.
It can provide direct and specific hint to generate image contents with the specified attributes \cite{reed2016generative}\cite{zhang2017stackgan}.
For example, for the first image in Figure 1b, a pretrained classification model can predict the image attribute vector to tell attributes such as ``wearing eyeglasses", ``male", ``not smiling", etc.
(2) Segmentation information is usually represented as a one-dimensional map that associates each pixel of an image with a region label, e.g., regions of eyes, mouth, and nose from a facial image.
It can tell the spatial relationship and boundaries among different segmentation regions \cite{Isola_2017_CVPR}\cite{wang2018high}, and therefore enforce the inpainting process to focus more on predicting and restoring object structures related to different segmentation regions.
Note that a wide range of large-scale image datasets labeled with attribute or segmentation information \cite{liu2015deep}\cite{zhou2017places}\cite{zhou2017scene}\cite{le2012interactive} are easily accessible nowadays.
Therefore, we aim to utilize and incorporate these two essential information in our inpainting process.
In order to effectively determine these two types of information for input images, we can pretrain state-of-the-art multi-label image classification models \cite{jin2016annotation}\cite{liu2017semantic}\cite{wang2016cnn}\cite{wei2016hcp}\cite{yang2016exploit} and image semantic segmentation models \cite{chen2018deeplab}\cite{chen2017rethinking}\cite{chen2018encoder}\cite{long2015fully}\cite{zhao2017pyramid} on auxiliary labeled datasets and incorporate them in image inpainting process.
The auxiliary labeled datasets do \emph{not} need to contain images that are exactly the same as those to be inpainted, as long as they are in similar categories.
For example, both images in our inpainting training dataset and its corresponding auxiliary labeled datasets are facial images, but they do not have to be images of the same person.

Based on this insight, we propose a novel \textit{boosted GAN with semantically interpretable information for image inpainting}.
Our proposed model is based on a GAN structure that consists of an inpainting network and a discriminative network.
The inpainting network employs two pretrained components: an attribute embedding network to predict attribute vectors of input images (images with missing regions) and a segmentation embedding network to predict segmentation maps of input images.
Then, we combine input images with their attribute vectors and segmentation maps to provide guidance for the generator to restore missing regions of the input images on both attribute and segmentation levels.
The generator follows an encoder-decoder structure with dilated convolutional layers used in the mid-layers to enlarge receptive fields \cite{iizuka2017globally}.
To examine whether a restored image is similar to its ground truth image not only on overall realness but also on attribute and segmentation levels, we propose a multi-level discriminative network.
It consists of three discriminators: a global discriminator that enforces overall realness, an attribute discriminator that regularizes attribute-level errors, and a segmentation discriminator that regularizes segmentation structure consistency with ground truth images.
As shown in Figure 1d, our model produces restored images with much better semantic consistency than those of the state-of-the-art model GntIpt \cite{yu2018generative}.
Our experimental results confirm that attribute regularization is essential for generating contents with correct attributes and sharp details, while segmentation regularization benefits alignment and consistency in segmentation structure.
We also conduct ablation study to analyze the effects of attribute regularization and segmentation regularization when used independently or combined with different trade-off parameters.
We find that the two regularizations have positive affects on each other and can achieve better results when incorporated together.
Besides, we introduced a novel semantic-level quantitative evaluation for image inpainting based on image retrieval.

In summary, the contributions of our paper are:
\begin{itemize}
  \item We propose a boosted GAN with semantically interpretable information for image inpainting that consists of an inpainting network and a discriminative network.
  \item The inpainting network utilizes an attribute embedding network and a segmentation network to discover the attribute and segmentation information of corrupted (input) images. 
  These two types of information are incorporated into the inpainting process to provide explicit semantic guidance.
  We also study the impact of inpainting quality contributed by the attribute information and segmentation information independently.
  \item The discriminative network adopts a multi-level design with three different discriminators.
  The global discriminator enforces overall realness with ground truth images, while the attribute discriminator regularizes attribute-level errors and the segmentation discriminator regularizes segmentation structure consistency.
  \item Experiments show that our model can effectively reduce the semantic inconsistency and generate visually plausible results compared with state-of-the art models. 
\end{itemize}

\section{Related Work}\label{secBackground}
\subsection{Generative Adversarial Networks}
The Generative Adversarial Network (GAN) \cite{goodfellow2014generative}\cite{Mirza2014Conditional}\cite{radford2016unsupervised} model is built on a game scenario consisting of two competitively learning networks: a generator and a discriminator.
The generator learns to generate samples that share similar distribution with the training data.
In contrast, the discriminator examines samples to identify whether they are from the generator's distribution or the training data distribution.
Both networks are trained alternatively and the competition drives them to improve until the generated samples are indistinguishable from the genuine samples.

\subsection{Image Inpainting}
Traditional image inpainting methods are mostly based on patch matching \cite{barnes2009patchmatch}\cite{bertalmio2003simultaneous} or texture synthesis \cite{efros1999texture}\cite{efros2001image}.
They suffer in the quality of the generated images when dealing with large arbitrary missing regions.
Recently, deep learning and GAN-based approaches have been employed to produce more promising inpainting results.
Phatak \textit{et al}. \cite{pathak2016context} first propose the \textit{Context Encoder} (CE) model that has an encoder-decoder CNN structure and train CE with the combination of reconstruction loss and adversarial loss \cite{goodfellow2014generative}.
The CE model can generate better images comparing with those generated by traditional methods.
However, the images generated by CE still tend to be blurry with evident artifacts and lack fine-grained details due to the limitation of the basic encoder-decoder generator structures.
Later methods use post-processing on top of images inpainted with encoder-decoder models to improve the quality of generate images.
For example, Yang \textit{et al}. \cite{Yang_2017_CVPR} take CE as the initial stage and refine inpainting results by propagating surrounding texture information.
Iizuka \textit{et al}. \cite{iizuka2017globally} make further improvement using global and local discriminators, followed by Poisson blending as post processing.
Yu \textit{et al}. \cite{yu2018generative} propose a refinement network based on contextual attention.
These methods mainly focus on enhancing the resolution of inpainted images, while ignoring the semantic consistency between the inpainted contents and the existing image context. 
Their models start with a classical encoder-decoder structure, which can easily suffer from the ill-posed one-to-many ambiguity when generating initial results.
Then the post-processing stage fails to take effect if given semantically incorrect intermediate results.
Liu \textit{et al}. \cite{liu2018image} propose partial convolution in order to utilize only valid pixels in convolution, which also fails to address the semantic inconsistency problem.

We refer to two previous studies \cite{yeh2017semantic}\cite{zhang2018semantic} that contain the phrase ``semantic image inpainting" in their titles.
Their notion of``semantic" is different from ours.
Specifically, they use ``semantic" to refer to the process of restoring the missing regions based on the context of input images (this is what GAN-based inpainting process does). 
In contrast, we explore explicit  human-understandable information (image attribute and segmentation information) from input images to boost and regularize the GAN-based inpainting process.

\subsection{Semantic Regularization for Deep Encoder-Decoders}  
Efforts have been made to improve autoencoder-based image synthesis models using semantic regularization.
Yan \textit{et al}. \cite{yan2016attribute2image} propose a conditional variational autoencoder with attribute-induced semantic regularization.
Reed \textit{et al}. \cite{reed2016generative} develop a generative adversarial model to generate images using text descriptions.
Zhang \textit{et al}. \cite{zhang2017stackgan} propose a two-stage process to further improve the synthesis results.
Our study differs from these studies in the following aspects: 
(1) We focuses on image inpainting, while these studies aim at text-to-image image synthesis.
(2) We introduce a unified semantic regularization strategy that can explore and integrate both attribute and segmentation information, while text-to-image synthesis tasks only use pre-labeled text description.
(3) Our inpainting training dataset does not need to be labeled and only images are provided as input during testing, while datasets for text-to-image image synthesis tasks are supposed to be labeled with corresponding text description. 

% !TEX encoding = UTF-8 Unicode
\section{Methodology}\label{secApproach}
Given an input image $x$ with missing regions (filled with zeros), image inpainting aims to restore the missing regions so that the output image $z$ can be consistent with the ground truth image $y$.
In this work, we utilize semantically interpretable information (attribute and segmentation information) of the input images (with missing regions) with the help of auxiliary labeled datasets.
Specifically, we set a \emph{prerequisite} for our model: auxiliary dataset $d_{1}$ labeled with attribute information and auxiliary dataset $d_{2}$ labeled with segmentation information should be available.
Images in these two datasets are supposed to have similar contents to those of the images in our inpainting training dataset $d_{0}$ (e.g., the images all contain human faces), but they do not have to be the same as those in $d_{0}$ (e.g., the images may be from different people).

\begin{figure*}[ht]
\vspace{-0.6cm}
\centering
\includegraphics[scale=0.55]{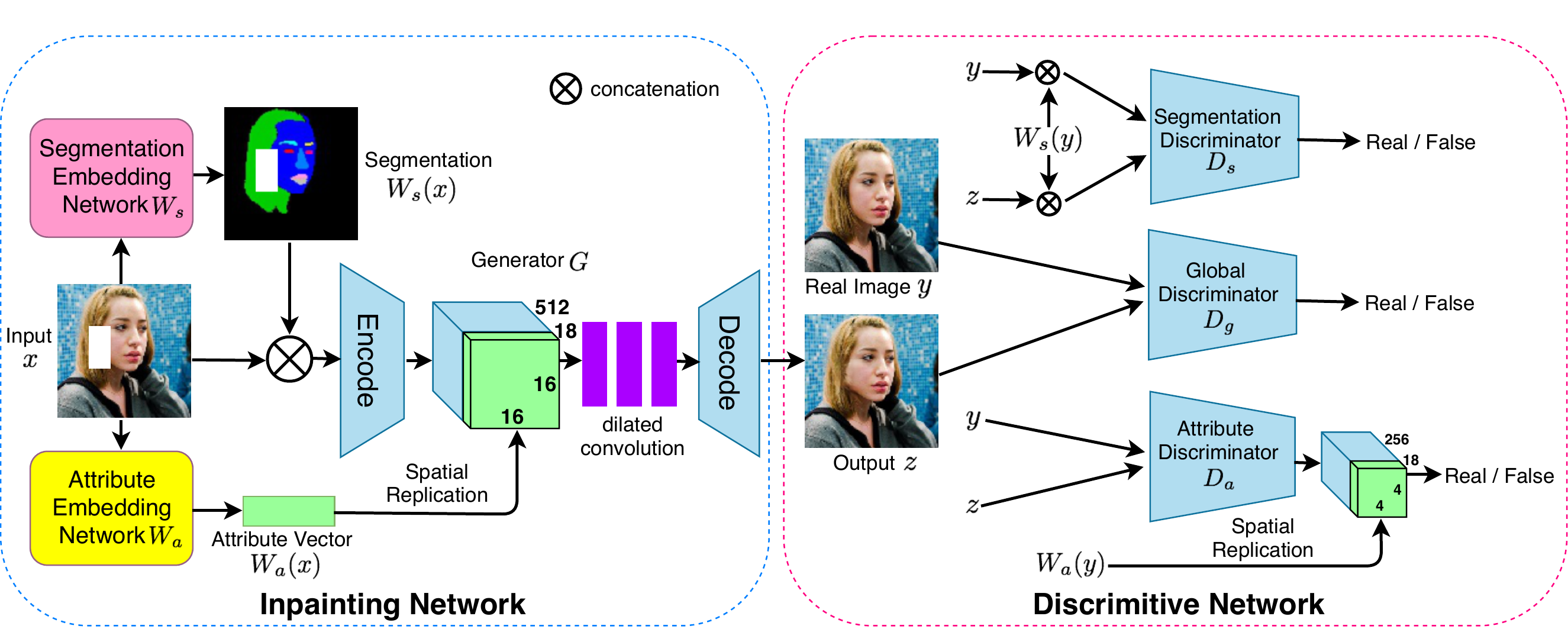}
\caption{Overview of our proposed model (best viewed in color).}
\label{structure}
\vspace{-0.7cm}
\end{figure*}

\subsection{Model Structure}
We propose a \textit{boosted GAN with semantically interpretable information for image inpainting}.
Our model is based on a GAN structure that consists of an inpainting network and a discriminative network.
The inpainting network (Section 3.1) utilizes two auxiliary embedding networks (pretrained on relevant auxiliary labeled datasets) to explore two types of semantically interpretable information (attribute information and segmentation information) of input images in order to provide explicit semantic-level guidance for the generator network.
The discriminative network (Section 3.2) focuses on constraining not only the output images to be similar to ground truth ones, but also the attribute and segmentation of output images to be consistent with those of the input images.
The architecture of our model is illustrated in Figure \ref{structure}.

\subsubsection{Inpainting Network}
The inpainting network consists of three sub-networks: an \textit{encoder-decoder generator network} $G$, an \textit{attribute embedding network} $W_{a}$, and a \textit{segmentation embedding network} $W_{s}$.
Our generator network is built on the architecture proposed by Iizuka \textit{et al}. \cite{iizuka2017globally}, which follows a general encoder-decoder structure, with dilated convolutional layers used in the mid-layers to enlarge receptive fields.
Since GAN-based image inpainting is inherently a one-to-many mapping problem, an autoencoder network alone may generate diverse contents that are in conflict with surrounding contexts. 
Unlike existing image inpainting methods that only take input images themselves as inputs, we propose to extract semantically interpretable information from these input images first and treat it as auxiliary information to boost inpainting process.
This strategy can provide explicit guidance on attribute level and segmentation level to increase semantic consistency between restored contents and surrounding contexts.
The following two networks are utilized to retrieve and embed the attribute and segmentation information with input images.

\textbf{Attribute Embedding Network}
Attribute information provides detailed text guidance for image inpainting.
The attribute embedding network is a multi-label classification network.
It aims to predict an attribute vector for each input (corrupted) image.
For example, in Figure \ref{structure}, the attribute vector of the input image predicted by the attribute embedding network can tell attributes such as 'female', 'young', 'oval face', \textit{etc}. 
We use the state-of-the-art multi-label classification model \cite{wang2017multi} as our attribute embedding network.
We pretrain this network on an external multi-label image dataset in which the images are similar to the ones in our inpainting training dataset.

\textbf{Segmentation Embedding Network}
Segmentation information tells the spatial relationship and boundaries among different segmentation regions.
The segmentation embedding network is a semantic segmentation network.
It aims to predict the segmentation maps for the input corrupted images.
We use the state-of-the-art semantic segmentation model \cite{chen2018encoder} as our attribute embedding network.
It is pretrained on an external image segmentation dataset in which the images are also similar to the ones in our inpainting training dataset.

To extract attribute and segmentation information, we first feed an input image $x$ into the attribute embedding network $W_{a}$ and the segmentation embedding network $W_{s}$ as shown in Figure \ref{structure}.
We obtain the attribute vector $W_{a}(x)$ and the segmentation map $W_{s}(x)$ of $x$ respectively.
The input image $x$ is concatenated with its segmentation map $W_{s}(x)$, and fed into the generator network $G$ which then becomes an intermediate feature map with a spatial size of $M_{1} \times M_{1}$.
Then the $N_{1}$ dimensional attribute vector $W_{a}(x)$ is spatially replicated to a $M_{1} \times M_{1} \times N_{1}$ tensor, and concatenated with the intermediate image feature map along the channel dimension as illustrated in the green and blue vectors of Figure \ref{structure}.
Finally, dilated convolutional layers and a series of upsampling layers (\emph{i.e.}, decoder) are used to generate the restored image $z$, which can be represented as follows:
\begin{equation}
z = G(x, W_{s}(x), W_{a}(x))
\end{equation}

\subsubsection{Discriminative Network}
The problem of semantic inconsistency poses a new challenge to the GAN discriminator: it now not only needs to measure the general similarity between restored images and real ones but also to provide explicit feedback for the generator to reach better consistency on both attribute and segmentation levels for the input and generated (restored) images. 
To address this challenge, we propose a multi-level discriminative network.
As illustrated in Figure~\ref{structure}, this network consists of three discriminators focusing on regularizations of different levels: a \textit{global discriminator} $D_{g}$, an \textit{attribute discriminator} $D_{a}$, and a \textit{segmentation discriminator} $D_{s}$.

The global discriminator examines the overall coherence between restored images and real images.
It is a binary classifier constructed by convolutional layers and fully connected layers to test whether its input image is fake (restored) or real.

The attribute discriminator first predicts the attribute vector $W_{a}(y)$ of the real image $y$ and then spatially replicates it to a $M_{2} \times M_{2} \times N_{1}$ tensor.
At the same time, the image (the generator output image $z$ or the real image $y$) is fed into the attribute discriminator to form a feature map with a spatial dimension of $M_{2} \times M_{2}$.
Then this feature map is concatenated with the attribute vector over the channel dimension and further fed to the remaining part of the attribute discriminator to obtain the decision score.

For the segmentation discriminator, both the restored image and the real image are concatenated with the real image's segmentation map $W_{s}(y)$ before being fed into the segmentation discriminator.
The structure of the segmentation discriminator is the same as that of the general discriminator.

We utilize the matching-based design \cite{reed2016generative} for both attribute discriminator and segmentation discriminator.
During training, the attribute (or segmentation) discriminator treats real images and their predicted attribute vectors (or segmentation maps) as positive sample pairs, while negative pairs have two groups: (1) real images and mismatched attribute vectors (or segmentation maps); (2) restored images and the predicted attribute vectors (or segmentation maps) of real images.

\subsection{Model Training}
Our model is trained in an end-to-end procedure.
In a single training epoch, the parameters of the discriminative network are first updated and then followed by those of the inpainting network, since the inpainting network depends on the back propagation of loss from the discriminative network. 
The objective function of the discriminative network $\mathcal{L}_{D}$ is:
\begin{equation}
\mathcal{L}_{D} = \mathcal{L}_{D_{g}} + \lambda_{a}\mathcal{L}_{D_{a}} + \lambda_{s}\mathcal{L}_{D_{s}}
\end{equation} 
where $\mathcal{L}_{D_{g}}$, $\mathcal{L}_{D_{a}}$, $\mathcal{L}_{D_{s}}$ represent the losses of the global discriminator, the attribute discriminator ,and the segmentation discriminator, respectively.
Tradeoff parameters $\lambda_{a}$ and $\lambda_{s}$ control the importance of the three terms.
The objective function of the global discriminator follows the original GAN loss function:
\begin{equation}
\mathcal{L}_{D_{g}} = -E_{y_{i} \sim P_{y}}[\log D_{g}(y_{i})] - E_{z_{i} \sim P_{z}}[\log(1 - D_{g}(z_{i}))]
\end{equation} 
where $z_{i}$ and $y_{i}$ represent a generated image and its corresponding real image;
$P_{y}$ and $P_{z}$ are the real and generated data distributions, respectively. 

For the attribute discriminator, we take the real image $y_{i}$ and its corresponding predicted attribute vector $W_{a}(y_{i})$ as a positive sample pair $\{y_{i}, W_{a}(y_{i})\}$.
Negative sample pairs consist of two situations: $\{z_{i}, W_{a}(y_{i})\}$ and $\{y_{i}, \overline{W}_{a}(y_{i})\}$, where $\overline{W}_{a}(y_{i})$ represents the mismatched attribute vector of $y_{i}$.
The objective function for the attribute discriminator is:
\begin{equation}
\begin{split}
\mathcal{L}_{D_{a}} = &-E_{y_{i} \sim P_{y}}[\log D_{a}(y_{i}, W_{a}(y_{i}))] \\
            &-E_{z_{i} \sim P_{z}}[\log(1 - D_{a}(z_{i}, W_{a}(y_{i})))] \\
            &-E_{y_{i} \sim P_{y}}[\log(1 - D_{a}(y_{i}, \overline{W}_{a}(y_{i})))]
\end{split}
\end{equation} 

Similarly for the segmentation discriminator, its objective function is formulated as follows:
\begin{equation}
\begin{split}
\mathcal{L}_{D_{s}} = &-E_{y_{i} \sim P_{y}}[\log D_{s}(y_{i}, W_{s}(y_{i}))] \\
            &-E_{z_{i} \sim P_{z}}[\log(1 - D_{s}(z_{i}, W_{s}(y_{i})))] \\
            &-E_{y_{i} \sim P_{y}}[\log(1 - D_{s}(y_{i}, \overline{W}_{s}(y_{i})))]
\end{split}
\end{equation} 
where $\overline{W}_{s}(y_{i})$ represents mismatched segmentation maps of real images $y_{i}$.

For our inpainting network, the adversarial loss contains three terms that provide guidance on overall similarity, attribute level, and segmentation level ,respectively.
Furthermore, we utilize $l_{2}$ distance as the reconstruction loss to minimize pixel-wise difference.
Hence, the object function $\mathcal{L}_{I}$ of the inpainting network is:
\begin{equation}
\begin{split}
\mathcal{L}_{I} = & E[\Vert z - y \Vert_{2} -\beta (\log D_{g}(z) + \lambda_{a}\log D_{a}(z, W_{a}(y))\\
            & + \lambda_{s}\log D_{s}(z, W_{s}(y)))] \\
      = & E_{(y_{i}, x_{i}) \sim P_{(y, x)}}[\Vert G(x_{i}, W_{s}(x_{i}), W_{a}(x_{i})) - y_{i}\Vert_{2} \\
        & - \beta \log D_{g}(G(x_{i}, W_{s}(x_{i}), W_{a}(x_{i}))) \\
        & - \beta \lambda_{a} \log D_{a}(G(x_{i}, W_{s}(x_{i}), W_{a}(x_{i})), W_{a}(y_{i})) \\
        & - \beta \lambda_{s} \log D_{s}(G(x_{i}, W_{s}(x_{i}), W_{a}(x_{i})), W_{s}(y_{i}))]
\end{split}
\end{equation} 
where $P_{(y, x)}$ indicates the joint distribution of the ground truth images and the corresponding input corrupted images.
$\beta$ controls the balance between the reconstruction loss and the adversarial loss.

% !TEX encoding = UTF-8 Unicode
\section{Experiments}\label{secExperiments}
To validate our method, both qualitative and quantitative evaluations are performed in this section.
We compare with four state-of-the-art inpainting methods:  \textbf{CE} \cite{pathak2016context}, \textbf{GL} \cite{iizuka2017globally}, \textbf{PConv} \cite{liu2018image}, and \textbf{GntIpt} \cite{yu2018generative}.
Following these studies \cite{iizuka2017globally}\cite{yu2018generative}, we also use CelebA-HQ \cite{karras2017progressive} and Places2 \cite{zhou2017places} (we select 20 scene categories) to train our model.
The images are resized and cropped to $256 \times 256$.

Before training our model on the inpainting dataset, we pretrain our attribute embedding network and segmentation embedding network on two auxiliary labeled datasets, respectively.
Both auxiliary datasets are supposed to have similar (not necessarily the same) contents  with the inpainting training dataset.
\textit{Note that the inpainting dataset for training our full model does not need to be labeled.}
For CelebA-HQ, we use CelebA \cite{liu2015deep} to pretrain the attribute embedding network (we select 18 attributes such as ``age", ``gender" out of a total of 40 attributes), and Helen Face dataset \cite{le2012interactive}\cite{smith2013exemplar}\cite{le2012interactive} to pretrain the segmentation embedding network.
For Places2, we use SUN \cite{xiao2010sun} to pretrain the attribute embedding network (we select 60 attributes: 20 scene categories that are the same as those of Places2 and 40 object categories), and ADE20k \cite{zhou2017scene} to pretrain the segmentation embedding network.

\begin{figure*}[t!]
\vspace{-0.1cm}
    \centering
    \setlength{\tabcolsep}{0.2em}
    {
        \begin{tabular}{ccccccc}
\includegraphics[width=2.1cm]{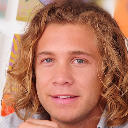}&
\includegraphics[width=2.1cm]{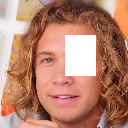}&
\includegraphics[width=2.1cm]{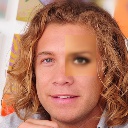}&
\includegraphics[width=2.1cm]{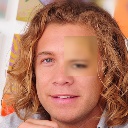}&
\includegraphics[width=2.1cm]{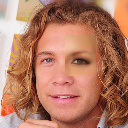}&
\includegraphics[width=2.1cm]{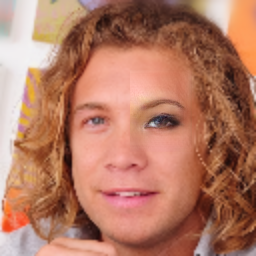}&
\includegraphics[width=2.1cm]{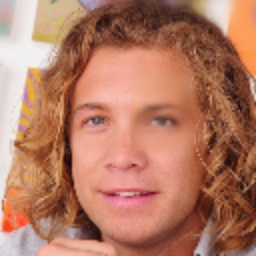}\\

\includegraphics[width=2.1cm]{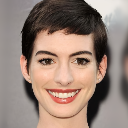}&
\includegraphics[width=2.1cm]{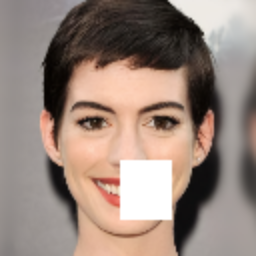}&
\includegraphics[width=2.1cm]{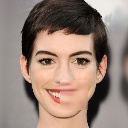}&
\includegraphics[width=2.1cm]{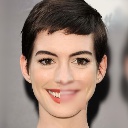}&
\includegraphics[width=2.1cm]{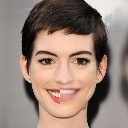}&
\includegraphics[width=2.1cm]{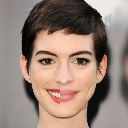}&
\includegraphics[width=2.1cm]{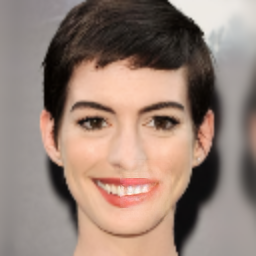}\\

\includegraphics[width=2.1cm]{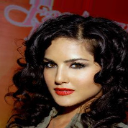}&
\includegraphics[width=2.1cm]{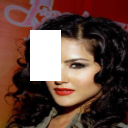}&
\includegraphics[width=2.1cm]{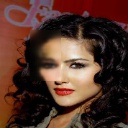}&
\includegraphics[width=2.1cm]{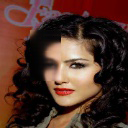}&
\includegraphics[width=2.1cm]{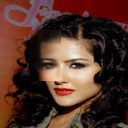}&
\includegraphics[width=2.1cm]{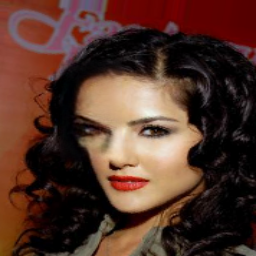}&
\includegraphics[width=2.1cm]{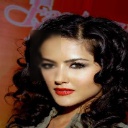}\\

(a) Original & (b) Input & (c) CE  & (d) GL  & (e) PConv  & (f) GntIpt & (g) Ours\\

\end{tabular}  
    
}   
    
    \caption{Qualitative comparisons on CelebA-HQ.}
\label{QualitativeCompare1}
\vspace{-0.5cm}
\end{figure*}

\textbf{Implementation Details}
By default, $M_{1} = 16$ and $M_{2} = 4$, while $N_{1}$ equals to the number of attributes we used to pretrain the attribute embedding network (18 for CelebA-HQ and 60 for Places2). 
Following existing work \cite{yu2018generative}\cite{iizuka2017globally}, we set $\beta$ as $0.01$.
Based on the analysis in Section \ref{ablative}, we empirically set $\lambda_{a} = 0.1$ and $\lambda_{s} = 0.1$.
We mask an image with a rectangular region that has a random location and a random size (ranging from $80 \times 80$ to $160 \times 160$).
The training takes one day on both CelebA \cite{liu2015deep} and Places2 \cite{zhou2017places} using an Nvidia GTX 1080Ti GPU.

\begin{figure*}[htb]
\centering
\setlength{\tabcolsep}{0.2em}
    {
\begin{tabular}{ccccccc}
\includegraphics[width=2.1cm]{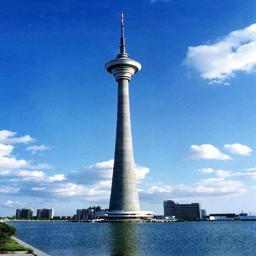}&
\includegraphics[width=2.1cm]{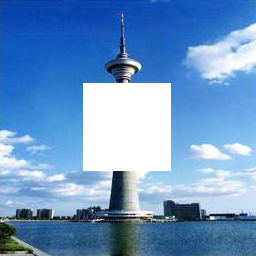}&
\includegraphics[width=2.1cm]{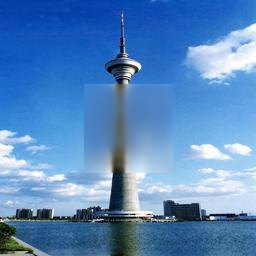}&
\includegraphics[width=2.1cm]{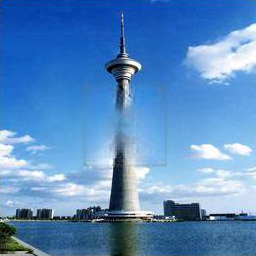}&
\includegraphics[width=2.1cm]{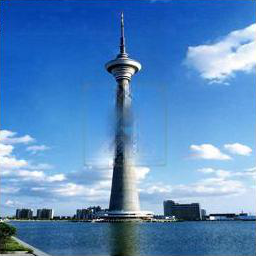}&
\includegraphics[width=2.1cm]{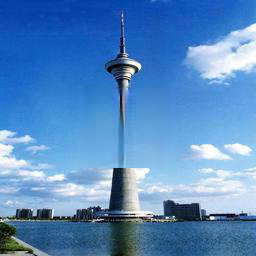}&
\includegraphics[width=2.1cm]{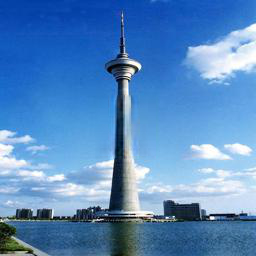}\\

\includegraphics[width=2.1cm]{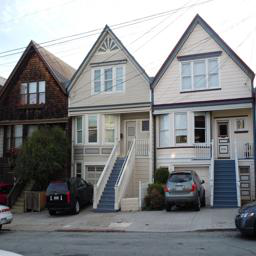}&
\includegraphics[width=2.1cm]{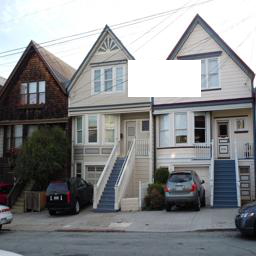}&
\includegraphics[width=2.1cm]{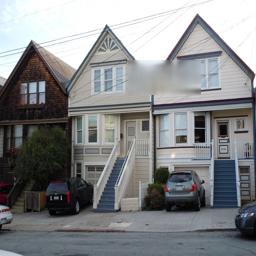}&
\includegraphics[width=2.1cm]{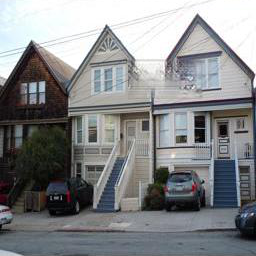}&
\includegraphics[width=2.1cm]{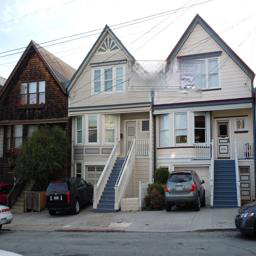}&
\includegraphics[width=2.1cm]{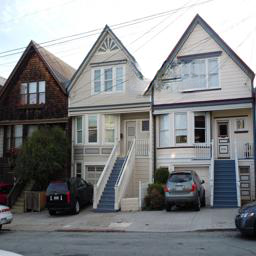}&
\includegraphics[width=2.1cm]{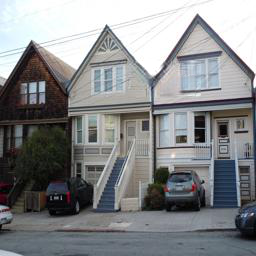}\\

\includegraphics[width=2.1cm]{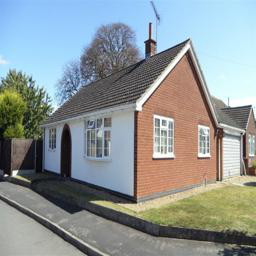}&
\includegraphics[width=2.1cm]{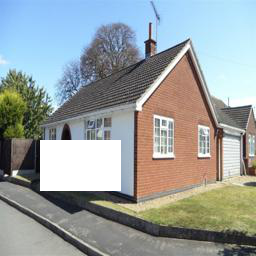}&
\includegraphics[width=2.1cm]{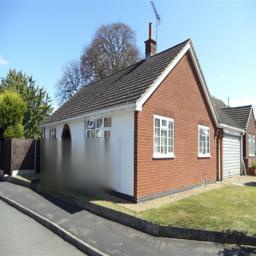}&
\includegraphics[width=2.1cm]{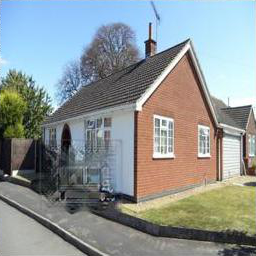}&
\includegraphics[width=2.1cm]{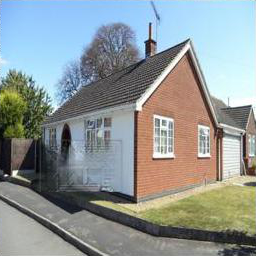}&
\includegraphics[width=2.1cm]{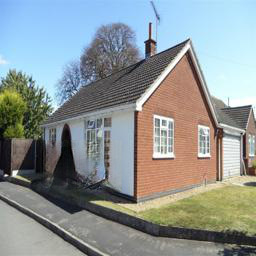}&
\includegraphics[width=2.1cm]{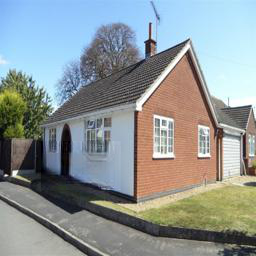}\\

(a) Original & (b) Input & (c) CE  & (d) GL  & (e) PConv  & (f) GntIpt  & (g) Ours\\

\end{tabular}
}
\caption{Qualitative comparisons on Places2.}
\label{QualitativeCompare2} % I can do without the label too
\vspace{-0.6cm}
\end{figure*}

\subsection{Qualitative Comparisons}
We show sample images in this section to highlight the performance difference between our methods and the four state-of-the-art methods on CelebA-HQ and Places2.
As shown in Figure \ref{QualitativeCompare1}, CE \cite{pathak2016context} can generate overall reasonable contents but the contents may be coarse and semantically inconsistent with surrounding contexts.
For example, the first row shows an image of a person that does not have heavy makeup. 
However, the generated content by CE is inconsistent on this attribute.
CE also encounters segmentation misalignment for the images in the second (misplaced mouth part) and third (blurry facial edge) rows.
This is due to the limitation of the original GAN model that only provides overall high-level guidance (whether the inpaining output is generally similar to the ground truth image) and lacks specific regularization in mid-level semantics such as attribute and segmentation.
GL \cite{iizuka2017globally} and PConv \cite{liu2018image} can generate better details than CE but still suffers from semantic inconsistency.
GntIpt \cite{yu2018generative} encounters attribute mismatch in the first (heavy make-up) and second (not smiling) rows and segmentation misalignment in the third (distorted facial edge) row.
Since the first stage of GntIpt resembles the overall structure of GL, if the coarse outputs from the first stage of GntIpt contain semantically inconsistent contents with the ground truth images, the second stage of GntIpt (\textit{i.e.}, refinement) does no help rescue the results and may even cause extra artifacts.
In contrast, our model can take advantage of the attribute and segmentation information that already exists in the input images, and can give explicit guidance on both attribute and segmentation level during inpainting process.
Therefore, our method can achieve better consistency over different attributes and better region boundary alignment compared with the four state-of-the-art methods.

We also evaluate the methods on Places2 as shown in Figure \ref{QualitativeCompare2}.
Again our method performs favorably in generating semantically plausible and photo-realistic images.

\subsection{Quantitative Comparisons}
Following previous studies \cite{yu2018generative}\cite{pathak2016context}\cite{Yang_2017_CVPR}\cite{yeh2017semantic}, we compute the mean $l_{1}$ error, mean $l_{2}$ error, \textit{Peak Signal-to-Noise Ratio} (PSNR) and \textit{Structural Similarity} (SSIM) over the restored image and the corresponding testing image for quantitative comparison.
As shown in Table \ref{tab0}, our proposed method outperforms the four methods among all the four metrics, and the results are consistent on both datasets.

\begin{table}[htb]
\caption{Quantitative comparisons on CelebA-HQ and Places2.}
\label{tab0}
\renewcommand\arraystretch{1}
\begin{center}
\scalebox{0.8}
{
\begin{tabular}{l|C|C|C|C}
      			\hline
      	 \multicolumn{5}{c}{CelebA-HQ}  \\ \hline
     			 Model & mean $l_{1}$ loss (smaller is better) & mean $l_{2}$ loss (smaller is better) & PSNR (larger is better) & SSIM \hspace{0.11cm} (larger is better)\\ \hline 
		CE  & 10.51\%  & 2.92\%  & 17.78 & 0.906\\	 
      		GL  & 9.60\%  & 2.57\%  & 18.09 & 0.923\\ 
      		PConv  & 9.33\% & 2.41\% & 18.61 &  0.938\\ 
      			GntIpt  & 9.18\% & 2.28\% & 18.80 &  0.940\\ \hline
      		\textbf{Ours} & \textbf{8.92\%} & \textbf{2.19\%} & \textbf{19.11} & \textbf{0.943}\\ \hline \hline
      		\multicolumn{5}{c}{Places2} \\ \hline
		CE  & 9.77\%  & 2.69\%  & 19.31 & 0.798\\	
      		GL  &  9.38\% & 2.21\% & 19.56 & 0.812\\ 
      		PConv    &  8.36\%  & 2.04\%  &  21.79 & 0.839\\
      			GntIpt    &  8.61\%  & 2.10\%  &  21.08 & 0.836\\ \hline
      		\textbf{Ours}  &  \textbf{8.26\%} & \textbf{1.97\%} & \textbf{22.32} & \textbf{0.845} \\ \hline
\end{tabular}    
}
\end{center}
\vspace{-0.7cm}
\end{table}

\begin{figure*}[htb]
\centering
\setlength{\tabcolsep}{0.2em}
    {
\begin{tabular}{cccccc}

\includegraphics[width=2.1cm]{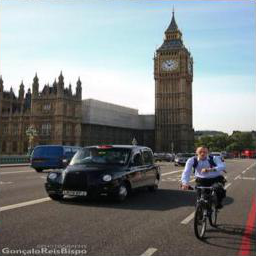}&
\includegraphics[width=2.1cm]{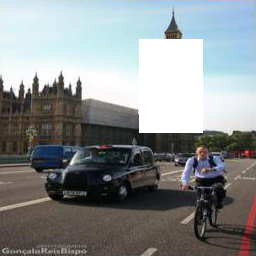}&
\includegraphics[width=2.1cm]{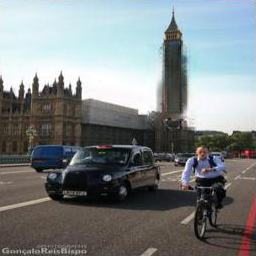}&
\includegraphics[width=2.1cm]{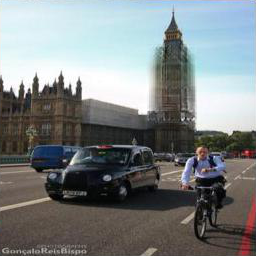}&
\includegraphics[width=2.1cm]{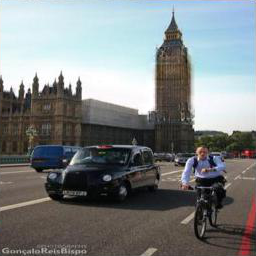}&
\includegraphics[width=2.1cm]{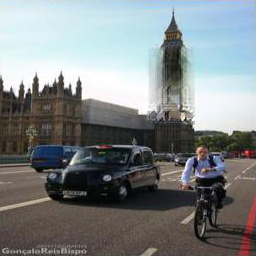}\\

(a) Original & (b) Input & (c) $\lambda_{a} = 0$ & (d) $\lambda_{a} = 0.01$  & (e) $\lambda_{a} = 0.1$ & (f) $\lambda_{a} = 1$\\

\end{tabular}
}
\caption{Qualitative comparisons of different $\lambda_{a}$ for attribute regularization. Here $\lambda_{s} = 0.1$.}
\label{ablative1} % I can do without the label too
\vspace{-0.5cm}
\end{figure*}

\begin{figure*}[htb]
\centering
\setlength{\tabcolsep}{0.2em}
    {
\begin{tabular}{cccccc}

\includegraphics[width=2.1cm]{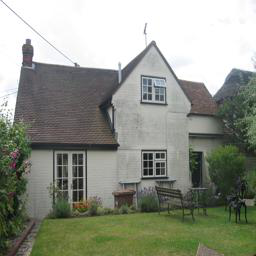}&
\includegraphics[width=2.1cm]{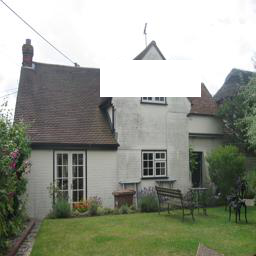}&
\includegraphics[width=2.1cm]{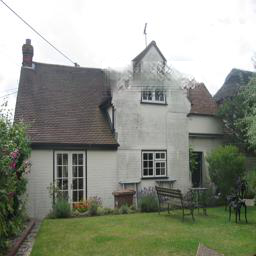}&
\includegraphics[width=2.1cm]{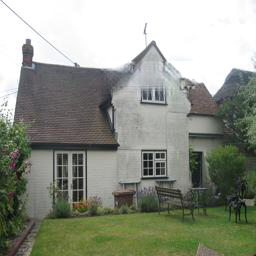}&
\includegraphics[width=2.1cm]{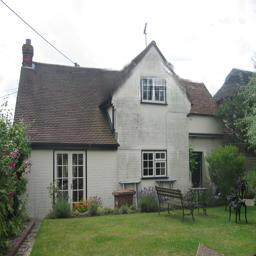}&
\includegraphics[width=2.1cm]{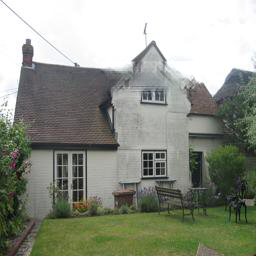}\\

(a) Original & (b) Input & (c) $\lambda_{s} = 0$ & (d) $\lambda_{s} = 0.01$  & (e) $\lambda_{s} = 0.1$ & (f) $\lambda_{s} = 1$\\

\end{tabular}
}
\caption{Qualitative comparisons of different $\lambda_{s}$ for segmentation regularization. Here $\lambda_{a} = 0.1$.}
\label{ablative2} % I can do without the label too
\vspace{-0.7cm}
\end{figure*}

\subsection{Ablation Study/Parameter Analysis}
\label{ablative}
We further conduct ablation experiments for our model on Places2.
Specifically, we investigate different combinations of attribute regularization and segmentation regularization in the loss function by (1) setting $\lambda_{a}$ as 0 and  varying $\lambda_{s} \in [0.01, 1]$, (2) setting $\lambda_{s}$ as 0 and  varying $\lambda_{a} \in [0.01, 1]$, (3) setting $\lambda_{a}$ as 0.1 and  varying $\lambda_{s} \in [0.01, 1]$, (4) setting $\lambda_{s}$ as 0.1 and  varying $\lambda_{a} \in [0.01, 1]$.
Qualitative comparisons are shown in Figure \ref{ablative1} and \ref{ablative2}.
Based on the visual results, we see that attribute regularization is able to capture attribute coherency and help generate sharp details, while segmentation regularization helps preserve region boundaries and structures.
%Specifically, we fix the tradeoff parameter $\lambda_{s}$ as 0.1 and train our model with variable values of tradeoff parameter $\lambda_{a}$.
%When $\lambda_{a}$ equals to zero which means only using segmentation regularization, the restored content is well-aligned with the surrounding segmentation structure, however almost no details can be observed.
%When $\lambda_{a}$ is small (\textit{e.g.}, $0.01$), the details of restored content is sharper but still blurry.
%When $\lambda_{a}$ becomes too large (\textit{e.g.}, $1$) which means the constrain is too excessive, extra artifacts and noises appear.
%Therefore, we empirically set $\lambda_{a} = 0.1$ in our experiments.
%The second group evaluates the effect of segmentation regularization as shown in Figure \ref{ablative2}.
%Accordingly we empirically set $\lambda_{s} = 0.1$ based on the comparisons.
Quantitative comparisons with 4 groups of parameter setting are given in Table \ref{tab1}.
We see that both attribute regularization and segmentation regularization contributes to the model performance when using them alone.
Moreover, the collaboration of these two different semantic regularizations in our model can achieve complementary effects to improve inpainting quality.
We empirically set $\lambda_{a} = 0.1$ and $\lambda_{s} = 0.1$ based on the ablation experiments.

\begin{table}[htb]
\caption{Quantitative comparisons with different combinations of $\lambda_{a}$ and $\lambda_{s}$ on Places2.}
\label{tab1}
\renewcommand\arraystretch{1}
\begin{center}
\scalebox{0.8}
{
\begin{tabular}{l|C|C|C}
      			\hline
      	 \multicolumn{4}{c}{$\lambda_{a} = 0$}  \\ \hline
     			 $\lambda_{s}$ & 0.01 & 0.1 & 1 \\ \hline 
		PSNR & 20.72  & \textbf{21.21}  & 20.38 \\	 
      		SSIM & 0.831  & \textbf{0.838}  & 0.832 \\ \hline \hline
      		
      		\multicolumn{4}{c}{$\lambda_{s} = 0$}  \\ \hline
     			 $\lambda_{a}$ & 0.01 & 0.1 & 1 \\ \hline 
		PSNR & 21.06  & \textbf{21.80}  & 21.54 \\	 
      		SSIM & 0.816  & \textbf{0.828}  & 0.820 \\ \hline \hline
      		
      		\multicolumn{4}{c}{$\lambda_{a} = 0.1$}  \\ \hline
     			 $\lambda_{s}$ & 0.01 & 0.1 & 1 \\ \hline 
		PSNR & 21.46  & \textbf{22.32}  & 21.84 \\	 
      		SSIM & 0.835  & \textbf{0.846}  & 0.840 \\ \hline \hline
      		
      		\multicolumn{4}{c}{$\lambda_{s} = 0.1$}  \\ \hline
     			 $\lambda_{a}$ & 0.01 & 0.1 & 1 \\ \hline 
		PSNR & 21.62  & \textbf{22.32}  & 22.15 \\	 
      		SSIM & 0.835  & \textbf{0.846}  & 0.838 \\ \hline
\end{tabular}    
}

\end{center}
\vspace{-0.7cm}
\end{table}

\subsection{Evaluation from Semantic Perspective}
Recent studies \cite{yu2018generative}\cite{Yang_2017_CVPR} find that inpainting results with high PSNR may turn out to be overly smooth and semantically unsatisfactory.
Thus, instead of relying only on low-level metrics such as PSNR, we propose a semantic-level metric based on image retrieval.
Specifically, given a specific image retrieval method $m$ and a image retrieval benchmark dataset, \textit{we first replace the ground truth images of each query image with the returned results of this query image by $m$}.
Then we put same masks on the original query images, and a certain inpainting method $i$ is implemented to recover these masked query images.
Finally we use the restored images as query images and conduct image retrieval on the benchmark dataset with the same image retrieval method $m$ to compute the mean average precision $mAP_{i}$, with respect to the retrieved results of original query images as our ground truth.
%Therefore, $mAP_{i}$ is the mean average precision of $m$ with the masked query images recovered by inpainting method $i$ as the input.
Therefore, $mAP_{i}$ can evaluate to what extent this inpainting method $i$ can narrow down the gap between the original query images and the restored ones, which directly measures how successfully the inpainting method $i$ restores the images from semantic level.
This can actually be done on any dataset since we do not need manual-labeled ground truth (we use the retrieved results of original query images as our ground truth).

\begin{table}[htb]

\renewcommand\arraystretch{1}
  \begin{center}
    \caption{Comparison of the proposed $mAP$ on benchmark datasets.}
    \label{tab4}
    \scalebox{1}
{
    \begin{tabular}{l|C|C|C}
      \hline
      Method & Paris & Oxford & Caltech256 \\ \hline \hline
      Masked & 76.23\% & 71.82\% & 84.28\% \\ \hline
      CE  & 83.82\% & 80.88\% & 89.73\% \\ 
      GL  & 88.79\% & 83.42\% & 92.06\% \\ 
      PConv  & 93.21\% & 87.27\% & 95.14\% \\ 
      GntIpt  & 91.65\%  &  87.36\% &  93.32\% \\ \hline
      %\textbf{Ours with only $W_{a}$} & \textbf{0.1347} & \textbf{0.1558} & \textbf{0.1026} \\
      %\textbf{Ours with only $W_{s}$} & \textbf{0.1262} & \textbf{0.1281} & \textbf{0.1103} \\
      \textbf{Ours} & \textbf{96.53\%} & \textbf{93.68\%} & \textbf{97.40\%} \\ \hline
    \end{tabular}
}
  \end{center}
  \vspace{-0.5cm}
 
\end{table}

To compare this proposed $mAP$ of the four methods and our method, we choose three image retrieval benchmark datasets (Paris \cite{philbin2007object}, Oxford \cite{philbin2008lost} and Caltech256 \cite{caltech}).
We resize the query images to $256\times256$ and employ a $128\times128$ missing region at the center of the images.
We use a state-of-the-art image retrieval method DF.FC2 in \cite{wan2014deep}, which uses the second fully connected layer of a pre-trained VGG-16 network \cite{Simonyan2014Very} to obtain the image feature representation and Euclidean distance to measure the similarity between query images and ground truth images.
Results of the $mAP$ values of the different models are shown in Table \ref{tab4}. 
Here, larger $mAP$ values are preferred, which means that using recovered images recovered from our model for image retrieval produces the most similar query results as those produced by using the original query images for image retrieval.
Our method outperforms the four state-of-the-art inpainting methods over all the three benchmark datasets.
Here, ``Masked" means the $mAP$ value of directly using the masked query images as input for image retrieval.
%Besides, it also provides quantitative proof that we successfully incorporate two different semantic regularization in our model and achieve complementary effects on inpainting. 

\section{Conclusion}\label{secConclusion}
We studied the semantic consistency problem in image inpainting and proposed a unified boosted GAN with semantically interpretable information for image inpainting that can generate contents consistent with the surrounding contexts both on attribute level and segmentation level.
The proposed inpainting network utilizes two auxiliary components: an attribute embedding network and a segmentation embedding network to discover the attribute and segmentation information of corrupted (input) images and incorporates them into the inpainting process.
The proposed multi-level discriminative network enforces regularizations not only on overall realness but also on attribute and segmentation consistency with the original images.
We evaluated the proposed method from both pixel level and semantic level.
The experimental results confirm that our proposed method outperforms the state-of-the-art methods consistently on real datasets.
Our model can be improved further, for example, since our attribute and segmentation embedding networks are pretrained on auxiliary datasets, attribute vectors and segmentation maps might not be accurately predicted for input images, and this might result in an inaccurate supervision for inpainting.
Our future work will explore domain-transfer-based strategies to improve the performance of our proposed method.

{\small
\bibliographystyle{ieeetr}
\bibliography{reference}
}

\end{document}